# A CNN-LSTM-BASED HYBRID DEEP LEARNING APPROACH TO DETECT SENTIMENT POLARITIES ON MONKEYPOX TWEETS

**Krishna Kumar Mohbey**
Department of Computer
Science, Central University
of Rajasthan Ajmer, India.

**Gaurav Meena**
Department of Computer
Science, Central University of
Rajasthan Ajmer, India.

**Sunil Kumar**
School of Business,
Woxsen University Hyderabad,
India

**K Lokesh**
Department of Computer Science
Central University of Rajasthan
Ajmer, India.

August 20, 2022

## ABSTRACT

People have recently begun communicating their thoughts and viewpoints through user-generated multimedia material on social networking websites. This information can be images, text, videos, or audio. Recent years have seen a rise in the frequency of occurrence of this pattern. Twitter is one of the most extensively utilized social media sites, and it is also one of the finest locations to get a sense of how people feel about events that are linked to the Monkeypox sickness. This is because tweets on Twitter are shortened and often updated, both of which contribute to the platform's character. The fundamental objective of this study is to get a deeper comprehension of the diverse range of reactions people have in response to the presence of this condition. This study focuses on finding out what individuals think about monkeypox illnesses, which presents a hybrid technique based on CNN and LSTM. We have considered all three possible polarities of a user's tweet: positive, negative, and neutral. An architecture built on CNN and LSTM is utilized to determine how accurate the prediction models are. The recommended model's accuracy was 94% on the monkeypox tweet dataset. Other performance metrics such as accuracy, recall, and F1-score were utilized to test our models and results in the most time and resource-effective manner. The findings are then compared to more traditional approaches to machine learning. The findings of this research contribute to an increased awareness of the monkeypox infection in the general population.

# 1 Introduction

The clinical manifestations of monkeypox are less severe than those of smallpox (a virus that spreads from animals to people). It manifests itself with symptoms that are similar to those of smallpox. Since smallpox was declared eradicated in 1980 and smallpox vaccinations were subsequently discontinued, monkeypox has taken the place of smallpox as the most significant orthopoxvirus about public health concerns. Monkeypox is a contagious disease that has been mostly afflicting people in the west and central Africa. However, it has recently expanded into urban areas and is often seen close to tropical rainforests. There are many different types of rodents and non-human primates that act as hosts for other creatures.

The monkeypox virus is a kind of the double-stranded DNA viral family Poxviridae, belonging to the genus Orthopoxvirus. Viruses of the monkeypox species have been traced back to both the Congo Basin and western Africa, suggesting that these two regions may be home to genetically distinct subgroups of the virus. Conventional wisdom held that the Congo Basin clade was the most infectious and caused the most severe illness. The boundaries of Cameroon serve as a de facto dividing line between the two viral clades since it is the only country in which both clades have been discovered. The monkeypox virus may be transmitted to humans from other animals, most often rats. Even while an epidemic may spread from person to person, it cannot be sustained indefinitely in this way. The signs and symptoms are similar to but not as severe as those of smallpox. The monkeypox virus is sometimes seen in the tropical rainforests of Central and West Africa, even though smallpox was declared extinct worldwide in 1980. The mortality rate during an outbreak of monkeypox has traditionally varied from 1 percent to 10 percent; however, most patients may completely recover with the correct treatment.

Social scientists and psychologists striving to understand better the human condition, psychology, and mental health are particularly interested in the exponential expansion of information that has happened in spreading through social media. Social media websites like Twitter have been used as a data collection tool for study in the behavioral science and psychology sectors. It has also been used to ascertain a user's personality type [1, 2] and learn about the patterns and histories of internet users.

With the help of sentiment analysis, a person's opinion may be automatically placed into one of three categories, depending on whether it is a positive opinion, adverse opinion, or neutral about a specific topic, product, movie, or news. NLP, the analysis of tweets, is used in this system to automatically analyze and quantify the users' feelings based on the information on Twitter. The management of positive, negative, and neutral sentiments communicated in tweets is the primary focus of the work being done on Sentiment Analysis at the moment. Analyzing sentiments is not difficult to access the opinions spoken in public assessments and survey replies. The recording of text data, the shooting of images, the creation of movies, and the identification of voice are only some of its possible applications. Following the data collection, the input is parsed into its parts, which may be individual words or phrases, for sentiment analysis, which operates on data comparable to that of tweets. Because of sentiment analysis on social media, which includes data from Twitter [3], there is a greater need for public perspectives, and these viewpoints are compiled in the form of text. It is also a challenging problem to accurately forecast data from tweets using text analysis to fulfill the requirements of commercial evaluation.

Twitter is a social networking service that has also become a significant outlet for disseminating media via the Internet. Lots of people rely on it and utilize it often. Real-time communication conveys information clearly and succinctly in relation to events as they occur, and it records people's opinions and reactions. Twitter is a social networking website where people may share their reactions to the global epidemic [4] regarding their feelings, thoughts, and opinions. People continue to use Twitter despite the current epidemic. However, the employment of ML-based algorithms is required due to the challenges in judging the inherent importance of a piece of content using NLP approaches, such as contextual phrases and words and ambiguity in written or spoken language [5-7]. NLP and its applications have significantly impacted text analysis and social media classification. Applications inspired by NLP have significantly affected the process of evaluating and classifying information found on social media.

Literally, "transferable learning" is the knowledge that may be used in other contexts. Transfer learning is often used in image processing to speed up the learning and training processes. Knowledge transfer is fundamental to transferring learning [8] from the source domain to the target domain. According to the study done by Sv et al. [22] the social media sentiment, more people are posting positively (28.82%) about the monkeypox virus than negatively (23.01%). A closer look at the tweets indicates that the bulk of the tweets with positive sentiments about monkeypox talk about the mild symptoms and lower infection mortality rate. A public opinion survey showed that people aren't too worried about the monkeypox virus. The analysis of tweets with negative sentiments about monkeypox revealed that users were discussing issues like the virus's potential to cause death, its severity, the lesions it leaves behind, the availability of vaccines, whether or not monkeypox is the next pandemic after COVID-19, the safety of travel, and the impact of the virus's spread on human health. Listed below is a breakdown of the technical aspects we contributed:

- This study finds the user opinion about Monkeypox infection on social media platforms like Twitter.
- The CNN-LSTM-based hybrid approach is proposed to detect the user's sentiments.
- Based on detected polarities, society may spread awareness of monkeypox infections.

After this section, the rest of the paper is organized as follows: first, a review of the relevant prior literature; second, an explanation of the methodology used for the various approaches; third, an account of the experiments conducted and the results obtained; and fourth, a discussion of the implications of the findings.

## 2   Related works

Due to its global impact on public health, monkeypox is of particular concern to countries in west and central Africa and the rest of the world. In 2003, the United States of America experienced the first country outside of Africa to have an outbreak of monkeypox. In this case, epidemiologists determined that exposure to ill prairie dogs in captivity was the source of the disease. These pets shared a cage with dormice and Gambian pouched rats that had been imported to Ghana. Over 70 cases of monkeypox have been documented in the US as a result of this outbreak. It has been reported that Nigerian tourists caught monkeypox in Israel in September 2018, the UK in September 2018, December 2019, May 2021, and May 2022, Singapore in May 2019, and the USA in July and November 2021. These nations may be found in Southeast Asia. Many cases of monkeypox were discovered in May of 2022 in countries where it was not ordinarily present. Studies are now being conducted to learn more about the disease's epidemiology, vectors, and transmission dynamics. [9]

The monkeypox virus is an orthopoxvirus that may infect humans and produce monkeypox, a viral illness with symptoms including fever and rash similar to smallpox. Since the smallpox virus was eradicated from the human population in 1980, monkeypox has become the most severe orthopoxvirus infection in humans. Cases are recorded most often from rural parts of nations located in Central and West Africa, in places near tropical rainforests where humans may have contact with animals that are afflicted with the disease. Someone can get monkeypox by coming into direct contact with the respiratory droplets of another person who has the disease, either at home or at a medical institution, or by coming into touch with contaminated objects or materials, such as bedding. Although these are the primary means of person-to-person transmission, monkeypox outbreaks often occur in small clusters of a few cases without progressing to extensive community transmission. This is because monkeypox is a very contagious disease. When swift action is taken in response to an epidemic, it is far simpler to contain the spread of the disease. Other instances of monkeypox are reported in different countries due to importation by travelers or animals afflicted by the disease [10].

The World Health Organization (WHO) convened an "emergency conference" [11] on May 20, 2022, to examine the worldwide worries over the increasing number of cases of the monkeypox virus. The meeting was called to address the global concerns. The WHO deliberated over the subsequent few days over whether or not the epidemic should be classified as a "potential public health emergency of international concern" (PHEIC), as was done in the past for outbreaks of COVID-19 and Ebola [12]. A "Level 2" monkeypox notice was issued by the Center for Disease Control (CDC) in the United States on June 6, 2022. This was in response to the significant rise in reported cases [13]. On July 23, 2022, the WHO proclaimed monkeypox as a global health emergency after another conference.

A generalized prediction model based on an artificial neural network (ANN) model was created by Y Kuvvetli et al. [14] to match the distributions of various nations and forecast the future number of daily cases and fatalities by COVID-19. They used data collected for many nations between March 11, 2020, and January 23, 2021. ANN model is another tool the government may use to help hospitals and other medical institutions avert problems.

Opinion mining (also known as emotional extraction) is a technique for analyzing customers' feelings and may be performed using either Text Mining (TM) or NLP. Advantages of adopting sentiment analysis include upselling potential, agent monitoring, and real-time data. Service providers may use sentiment analysis to figure out what customers are happy with and what they aren't [16]. Discerning positive and negative sentiments in complex wordplays may be challenging, but sentiment analysis can help with that [17]. Using the sentiment analysis technique, researchers can determine the overall tone of a piece of writing. Information gathering is improved and made more thorough via sentiment analysis.

It was proposed by Hassan et al. [18] to identify polarity by combining CNN and LSTM with pre-trained word vectors obtained from IMDB movie reviews. The four-layer CNN classifier that was developed using this strategy consisted of two levels of convolutional processing, two tiers of pooling processing, and two layers of output processing. Both the CNN model, which had an accuracy of 87.0%, and the LSTM model, which had an accuracy of 81.8%, fared worse in the trials than the combined CNN + LSTM model, which had an accuracy of 88.3%. Shen et al. [19] developed a unique approach for determining the polarity of movie reviews by integrating CNN with bidirectional LSTM. This method successfully recognized both positive and negative reviews. When combined, the CNN and LSTM classifiers produce a model with an accuracy of 89.7%, which is much higher than the individual models' accuracy of 83.9 and 78.5%, respectively. The four-layer CNN classifier that was developed using this strategy consisted of two levels of convolutional processing, two tiers of pooling processing, and two layers of output processing. When the CNN and LSTM models were used jointly, the accuracy climbed to 88.3% when used alone from 81.8% in the trials.

Short-term memory is a significant obstacle for most RNNs, making LSTMs an essential machine learning component. Because of this, it becomes difficult to retain data from one process and apply it to another. For instance, if a dataset or numbers are projected to be processed, Recurring Neural Networks may forget some historical information [22]. The investigation of human-related occurrences has been significantly aided by the processing and analysis of data obtained from social media platforms, which has transformed the field of infodemiology. In addition, these social networks publish various statistical data on social-trend illnesses like monkeypox, such as the number of comments, images, and videos that have been shared. As a result, this makes it possible to anticipate the monkeypox morbidity rates in each region and alerts the decision-makers in charge of health policy to the need to develop educational and preventative initiatives in the areas with the highest risk [23].

## 3  Sentiment polarity detection using the CNN-LSTM hybrid approach

The dataset used in this study was sourced from the open-source hosting site GitHub. All sorts of tweets about monkeypox are included in this collection. After finishing the pre-processing and making sure there were no duplicate or null values, the data characters were detokenized so that the sentences could be split down into words and labels could be assigned. After that, the performance of the forecasting model was evaluated using deep learning's CNN-LSTM architectures. Figure 1 is a detailed schematic of the whole system I made using the methods I just outlined.

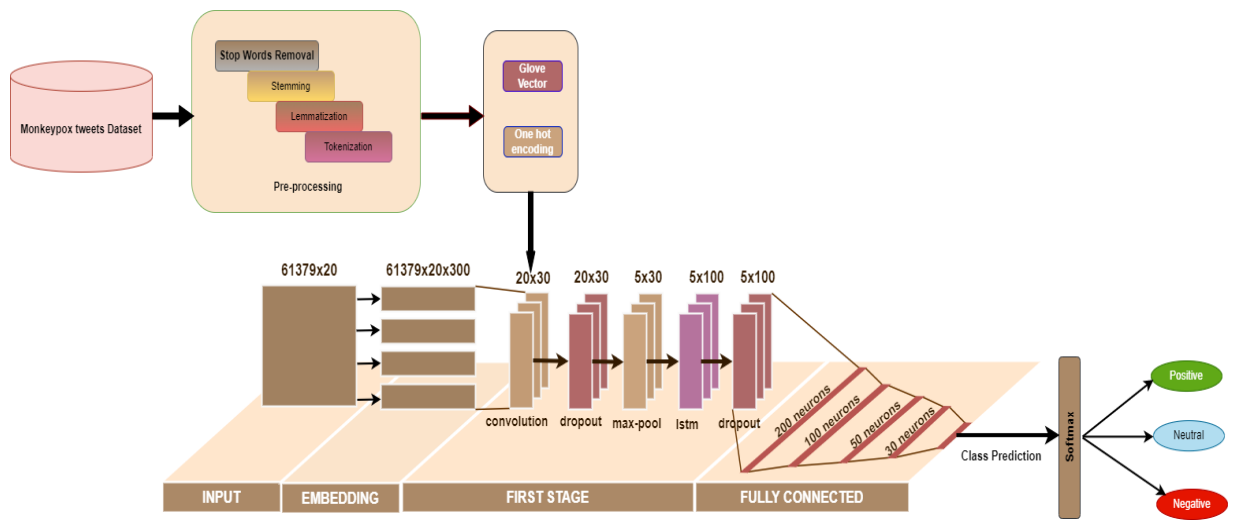

Fig.1. Outline of sentiment analysis procedures

### 3.1 CNN-LSTM

The architecture for sentiment categorization is described in this part and will be applied to the Twitter data analysis. We used an architecture based on recurrent neural networks termed CNN-LSTM to test the models. The CNN-LSTM architecture is seen in figure 2.

`

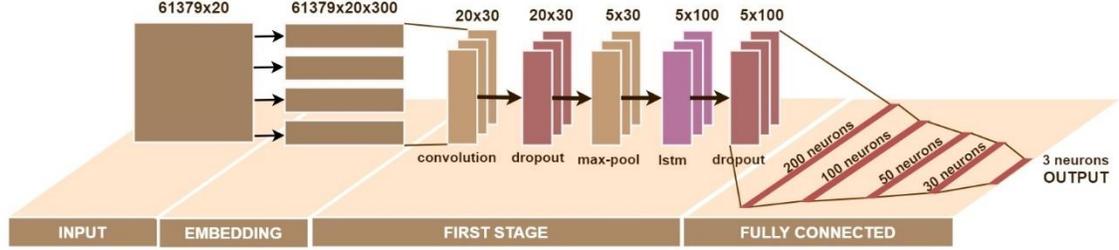

Fig.2. CNN-LSTM architecture for proposed sentiment analysis

Assume that a string represents the word vectors and that $P_i \in R^k$ represents the K-dimensional vector comparable to the $i^{th}$ token in a user review of total size n, where n is the number of tokens in the Equation (1). Zero padding is added if the sentence is smaller than n characters.

$$P_{1:n} = P_1 + P_2 + P_3 . \ldots \ldots \ldots + P_n \qquad (1)$$

The + operator denotes a concatenation operation in Equation (1). Similarly, suppose the concatenation of the words $P_i$ , $P_{i+1}$ , $P_{i+2}$ , . . . . . . . . . $P_{i:i+j}$ is equivalent to $P_{i:i+j}$ . Let $W \in R^{hk}$ denote the convolutional filters applied in an nxK dimensional matrix of a sentence with a window or gap of h words to produce a new feature matrix. The basic element $P_{i:i+j}$ denotes the local feature matrix from the ith to the $(I + J)^{th}$ line of the present sentence vector. Equation (2) can be used to produce a feature $C_{if}$ from a window of words $P_{i:i+h-1}$ .

$$C_i = f (W \cdot P_{i:i+h-1} + b) \qquad (2)$$

An activation function, such as a hyperbolic tangent or sigmoid, is used at the place where the bias, denoted by b, is one of the real numbers. The actual numbers include b as one of their members. In order to generate a feature map through Equation (3), the convolved convolutional filter needs to be applied to every window containing words.

$$C = [C1, C2, C3, C_{n-h+1}] \qquad (3)$$

Where C belongs to $R^{n-h+1}$.

The Equation for building a single feature map from a single convolutional filter is shown above and may be thought of as follows: In the same manner, m(n-h + 1) features will be generated by a convolutional layer that has multiple m filters. On feature maps, the max pooling layer is not used because feature selection might create disruptions to long-term dependencies early on in the LSTM layers. In order to adequately capture long-term dependencies, the features are immediately moved into the LSTM layer, which comes before the fully connected layer.

### 3.2 Pre-processing

Before actual tests on the gathered tweets, pre-processing is an essential step that must first be completed. The collected tweets are chaotic, unbalanced, and contain many stop words. Therefore, to do classification and prediction tasks, it is necessary to clean all of the tweets. Since Twitter is an unstructured platform that permits publishing in several languages, data pre-processing is of utmost significance. For this study, we restricted ourselves to collecting only tweets written in English. The removal of stop words, stemming of words, data filtering, and feature extractions are all included in pre-processing [24].

    *A. Stop word removal*
    We have stripped each tweet of any and all used terms during this stage. A list of stop words that had been predefined was utilized in the process of deleting stop words. Each tweet is checked against the available stop word list, and any terms that appear on both lists are eliminated from the respective tweets. These terms do not provide any positive contributions to the model's functioning.

    *B. Punctuation removal*
    Because the vast majority of the punctuation that may be used in tweets is meaningless, this step removes all of the punctuation symbols, including,!, ^, ?,.

    *C. URL removal*
    The following step is to clear out the tweets of all URLs and hyperlinks.

    *D. Filtering*
    Identify and remove hashtags, RT (retweet symbols), query phrases, and special characters.

    *E. Stemming*
    Porter Stemmer [23] used stemming from each tweet's phrases. All the terms are changed to the root word throughout this procedure.

## 4 Experimental results

### 4.1 Data source and description

Tweets are thought to be associated with monkeypox in this particular piece of research. The data set includes 61379 tweets concerning monkeypox that were published on Twitter on May 7 and June 11, 2022 [25]. This dataset includes a variety of tweets posted by individuals, some of which express the users' positive thoughts, some of which express their negative ideas, and some of which express their neutral viewpoints. Figure 3 illustrates the classification of tweets according to the categories to which they belong. In figure 3, the 0 class represents those with neutral opinions, the -1 class represents those with negative thoughts, and the 1 class represents those with positive opinions.

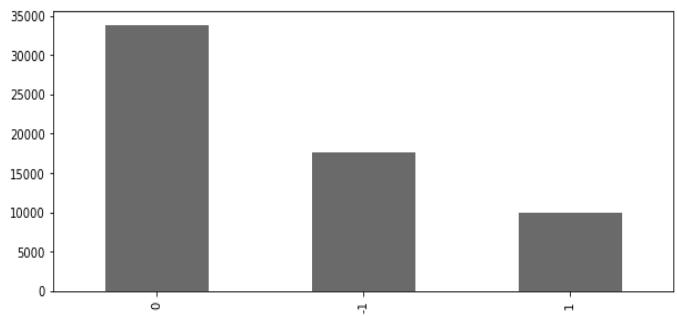

Fig. 3. The histogram of various classes of Monkeypox tweets dataset

### 4.2 Experimental Setting

Python is what we use on a machine running X86-64 Ubuntu 18.04.4 LTS. The CPU is an Intel(R) Core(TM) i7-8550U operating at 1.80GHz, with 16 gigabytes of RAM. We use this setup to investigate sentiment emotions with the suggested CNN-LSTM model. In order to recognize the characteristics, the CNN architecture was given training. Following that, other characteristics from the LSTM architecture were utilized to classify people's feelings.

### 4.3 Performance evaluation

The correct evaluation procedures are required to eliminate the bias present in the models' differentiation. Precision (P), recall (R), F1 score, accuracy, and AUC are some of the metrics that are used for the classification standard the most commonly [26, 27]. [P] stands for precision, recall (R) for recall and accuracy, and AUC stand for the area under the curve. The fraction of properly recognized samples relative to the total number of samples that have been evaluated and identified is used to determine how accurate the results are. The recall is the total number of samples from all positive representations that were successfully identified. This figure may be broken down further into individual samples. The F1 score is arrived at by taking the recall and accuracy scores and averaging them harmonically. The accuracy may be determined by calculating the proportion of correctly recognized samples that are included within the total number of samples [28]. The AUC is a statistic that may be used to describe the ROC curve [29]. It measures how well a classifier can differentiate between several different data groups. Table 1 presents the mathematical expressions of various measures.

Table 1: Evaluation expressions

| Measure | Expression | Measure | Expression |
|---|---|---|---|
| $P$ | $\dfrac{TP}{TP + FP}$ | $F1\ Measure$ | $\dfrac{2 * P * R}{P + R}$ |
| $R$ | $\dfrac{TP}{TP + FN}$ | $AUC$ | $\dfrac{\left(R - \dfrac{FP}{FP + TN} + 1\right)}{2}$ |
| $Accuracy$ | $\dfrac{TP + TN}{TP + TN + FP + FN}$ | | |

In Table 3, TN refers to the number of negative samples that were correctly identified, while TP refers to the number of positive samples that were correctly identified. In statistics, FP is the number of times unfavorable instances were incorrectly recognised as positive, while an FN predicted value is negative, but the actual value is positive. Table 2 contains an exhaustive listing of the confusion matrix's parameters.

Table 2: Confusion matrix

| | | Actual Value | |
|---|---|---|---|
| | | Positive | Negative |
| Predicted Value | Positive | TP | FP |
| | Negative | FN | TN |

## 4.4 Experimental outcomes

In order to determine whether or not the proposed CNN-LSTM model is valuable, we compared the outcomes of the experiments to the findings obtained from various machine learning models. This allowed us to determine how efficient the proposed model is. During the comparison, each of the following was considered: accuracy, the area under the curve, f1-score, precision, recall, and confusion matrix. The model that has been offered and is now being discussed uses various classifications, including neutral, positive, and negative categories. Following each epoch of training for the monkeypox tweet dataset, figures 4, 5, and 6 demonstrate the accuracy and loss history on several deep learning-based models' test sets.

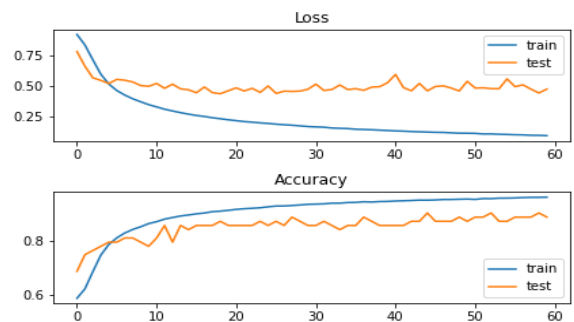

Fig.4. The *accuracy* and *loss history* of the CNN model

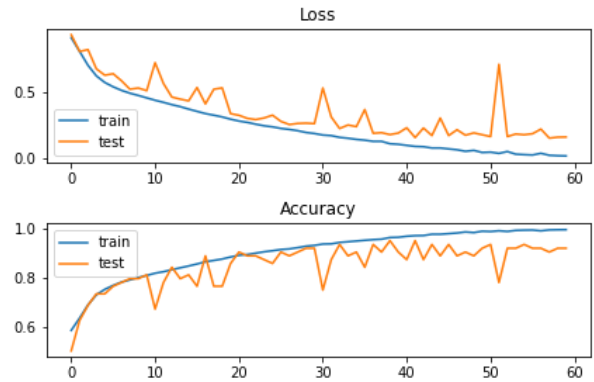

Fig. 5. The *accuracy* and *loss history* of the LSTM model

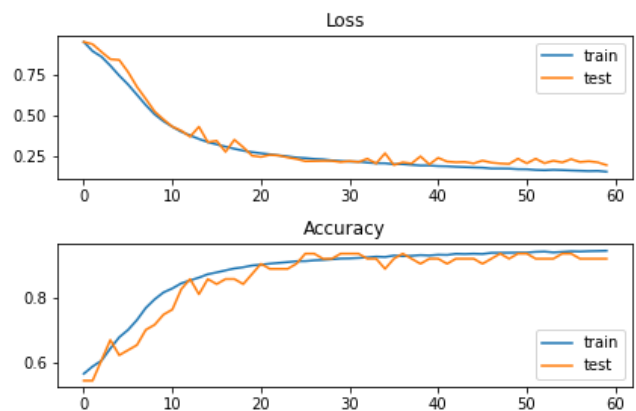

Fig. 6. The *accuracy* and *loss history* of the CNN-LSTM model

It is evident from the data that the CNN-LSTM model outperforms more straightforward CNN and LSTM-based deep learning models in terms of performance. In figure 8, accuracy increases steadily from epoch 10 and remains constant until epoch 50, at about 94%. It will be regarded as the experiment's most acceptable outcome. Eight models' findings from the monkeypox tweets dataset are displayed in Figure 7. In terms of Accuracy, Precision, Recall, F1-score, and AUC across all five assessment parameters, the suggested technique performs better than existing state-of-the-art methods overall. After epoch 60, the loss for the CNN-LSTM is roughly 0.1, and accuracy can approach 94%. This shows that our suggested model has more accuracy, indicating that it can accurately assess different moods. Bold is used to emphasize the best results.

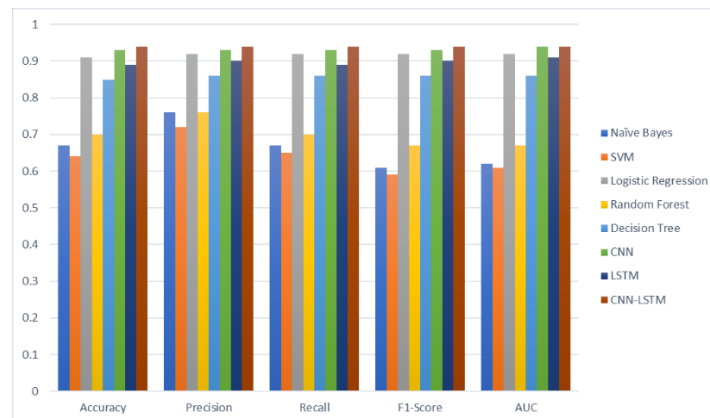

Fig. 7. The performance comparison of different models

Figure 8 displays the confusion matrices generated on the hidden test dataset to help the reader better understand the classification performance of each model. The CNN-LSTM model has the highest classification accuracy, as seen in figure 7.

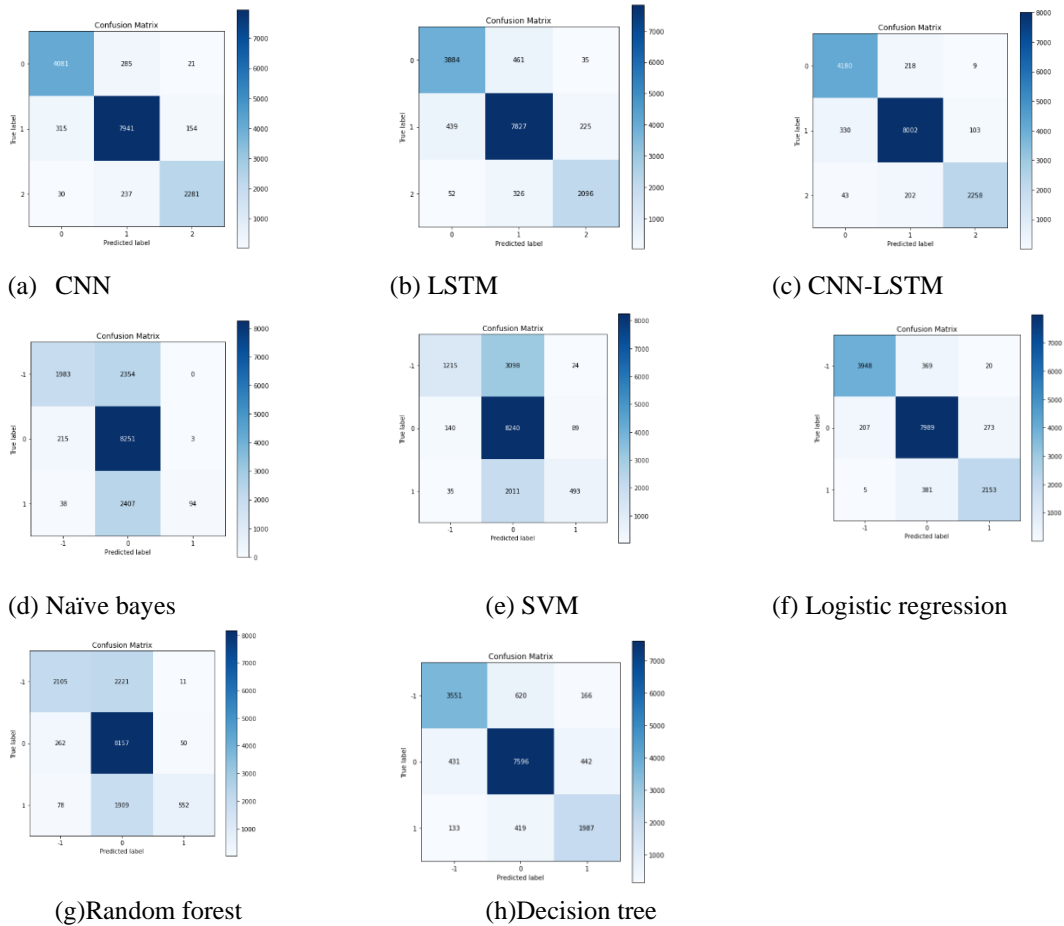

(a) CNN
(b) LSTM
(c) CNN-LSTM

(d) Naïve bayes
(e) SVM
(f) Logistic regression

(g)Random forest
(h)Decision tree

Fig. 8. The obtained confusion matrix of different models

## 4.5 Discussion

The results produced by the CNN- LSTM model that was built are compared to the findings produced by a variety of other models that are regarded as being state-of-the-art. In addition to such models, this collection also contains models that use machine learning. The research findings indicate that when compared to the other models, the CNN-LSTM model yields results of a higher quality than those produced by the other models. When applied to the dataset of tweets on monkeypox, the f1-score measure that the LSTM model generates is correct 94% of the time. In addition to this, the accuracy of the recommended model, and its recall, have both been greatly improved.

## 5 Conclusion

The findings of our study illustrate how CNN and LSTM techniques may be utilized to analyse tweets to determine the emotional polarity of their contents. The results of this study are based on the user's perception of the monkeypox infectious sickness as either positive, negative, or neutral. The CNN-LSTM model we suggested as part of our study helped us identify how precisely and effectively, we could anticipate and assess the sentiments of people on social media platforms like Twitter. The new model demonstrated superior performance compared to the methods already in use. In the case of the monkeypox datasets, the hybrid CNN-LSTM architecture with hyperparameter tunings achieved a level of accuracy of 94%. According to the findings, the proposed technique is successful and suitable for categorizing the sentiments in tweets about monkeypox. In addition, the proposed work is helpful for society because it raises public awareness of the recently emerged virus monkeypox. An iterative method that employs the more robust techniques, which may be refined further in subsequent research, may be used to optimize the characteristics of a system. This optimization can be accomplished. It is also conceivable to modify it to work, so that subject identification and sentiment classification may be carried out simultaneously. It would be helpful in system improvement.